\documentclass[10pt,twocolumn,letterpaper]{article}
\usepackage{wacv}
\usepackage{times}
\usepackage{epsfig}
\usepackage{graphicx}
\usepackage{amsmath}
\usepackage{amssymb}
\usepackage{algpseudocode}
\usepackage{algorithm}
\usepackage{amsfonts}
\usepackage{floatflt}
\usepackage{caption}
\usepackage{subcaption}
\usepackage[table,xcdraw]{xcolor}
\usepackage{microtype}
\usepackage[clock]{ifsym}
\usepackage{booktabs}
\usepackage[pagebackref=true,breaklinks=true,letterpaper=true,colorlinks,bookmarks=false]{hyperref}
\wacvfinalcopy 

\def\wacvPaperID{307} 

\ifwacvfinal\fi
\definecolor{marioscomment}{RGB}{255,0,255}

\newcommand{\pose}{\boldsymbol{\theta}}
\newcommand{\shape}{\boldsymbol{\beta}}
\author{
Hosnieh Sattar$^{1}$
\and
Gerard Pons-Moll$^1$\\
\and
Mario Fritz$^2$
\and
$^1$Max Planck Institute for Informatics, $^2$CISPA Helmholtz Center i.G.\\
Saarland Informatics Campus, Saarbr\"ucken, Germany, \\
{\tt\small \{sattar,gpons\}@mpi-inf.mpg.de, fritz@cispa.saarland}
}

\begin{document}

\title{Fashion is Taking Shape: \\Understanding Clothing Preference Based on Body Shape From Online Sources}

\maketitle
\ifwacvfinal\thispagestyle{empty}\fi

\begin{abstract}
To study the correlation between clothing garments and body shape, we collected a new dataset (Fashion Takes Shape), which includes images of female users with clothing category annotations. Despite the progress in body shape estimation from images, it turns out to be challenging to infer body shape from such diverse, real-world photos. Hence, we propose a novel and robust multi-photo approach to estimate body shapes of each user and build a conditional model of clothing categories given body-shape. We  demonstrate that in real-world data, clothing categories and body-shapes are correlated and show that our multi-photo approach leads to a better predictive model for clothing categories compared to models based
on single-view shape estimates or manually annotated body types.
We see our method as the first step towards the large-scale understanding of clothing preferences from body shape.
\end{abstract}

\section{Introduction}

Fashion is a $\$2.4~$ trillion industry\footnote{https://www.mckinsey.com/industries/retail/our-insights/the-state-of-fashion} which plays a crucial role in the global economy. Many e-commerce companies such as Amazon or Zalando makes it possible for their users to buy clothing online. However, based on a recent study, \footnote{https://www.ibi.de/files/Competence\%20Center/Ebusiness/PM-Retourenmanagement-im-Online-Handel.pdf} around $50\%$ of bought items were returned by users. One major reason of return is "It doesn’t fit" $(52 \%)$.  Fit goes beyond the mere size --- certain items look good on certain body shapes and others do not.
In contrast to in store shopping where one can try on clothing, in online shopping users are limited to a coarse set of numeric size ranges (e.g. 36, 38 and so on) to predict the fitness of the clothing item. 
Also, they only see the clothing worn by a professional model, which does not account for the diverse body shape of  people. 
A clothing item that looks very good on a professional model body could look very different on another person's body. 
\begin{figure}[!t]
\centering
\includegraphics[width=0.70\linewidth]{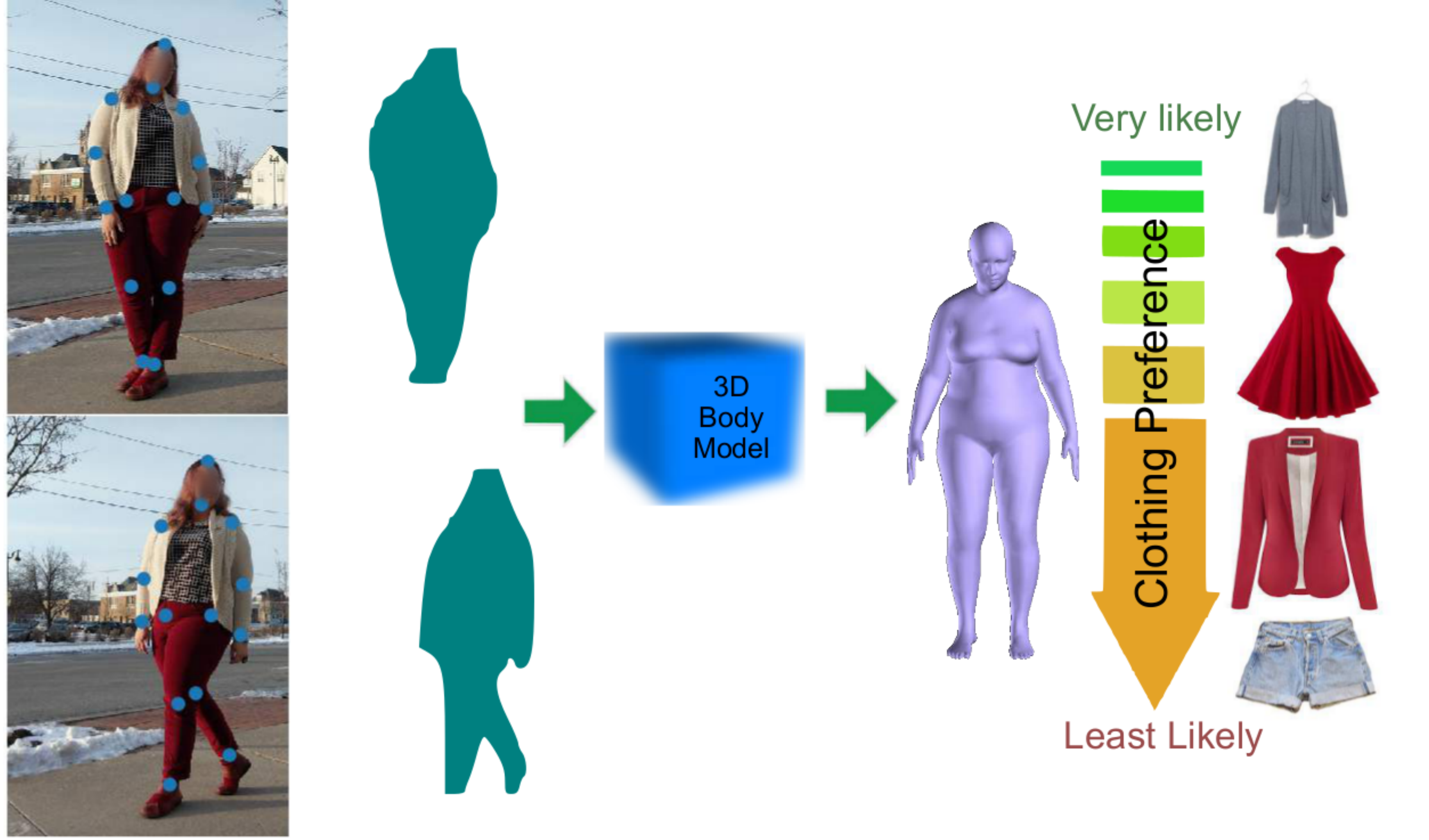}
\caption{Our multi-photo approach uses 2D body joint and silhouette to estimate 3D body shape of the person in the photo. Our shape conditioned model of clothing categories uses the estimated shape to predict the best fitting clothing categories.}\label{fig:teaser}
\end{figure}
Consequently, understanding how body shape correlates with people's clothing preferences could avoid such confusions and reduce the number of returns. 

Due to importance of the fashion industry, the application of computer vision in fashion is rising rapidly. Especially, clothing recommendation ~\cite{HipsterWarsECCV14,SimoSerraCVPR2015,liuLQWTcvpr16DeepFashion,han2017learning} is one of the hot topics in this field along with cloth parsing~\cite{6248101,yang2014clothing,gong2017look}, recognition~\cite{chen2012describing,liuLQWTcvpr16DeepFashion,Hsiao_2017_ICCV,han2017automatic,8237312} and retrieval~\cite{Wang:2011:CSC:2072298.2072013,6751549,kiapour2015buy,Liu2012StreettoshopCC}. 
Research in the domain of clothing recommendation studies the relation between clothing and categories,  location, travel destination and weather. 
However, there is no study on the correlation between human body shape and clothing. This is probably due to the fact that there exists no dataset with clothing category annotations together with detailed shape annotations. 

Therefore, our main idea is to leverage fashion photos of users including clothing category meta-data and, for every user, automatically estimating their body shape. Using this data we learn a conditional model of clothing given the inferred body shape.   

Despite recent progress, the visual inference of body shape in unconstrained images remains a very challenging problem in computer vision. People appear in different poses, wearing many different types of garments, and photos are taken from different camera viewpoints and focal lengths.

Our key observation is that users typically post several photos of themselves, while viewpoint and body pose varies across photos, the body shape \emph{does not} (within a  posted gallery of images). 
Hence, we propose a method that leverages multiple photos of the same person to estimate their body shape. Concretely, we first estimate body shape by fitting the SMPL body model~\cite{SMPL:2015} to each of the photos separately, and demonstrate that exhaustively searching for depth improves performance. Then, we reject photos that produce outlier shapes and optimize for a single shape that is consistent with each of the inlier photos. This results in a robust \emph{multi-photo} method to estimate body shape from unconstrained photos on the internet.

We use image from the web \footnote{http://www.chictopia.com} and collected a dataset (\emph{Fashion Takes Shape}) which includes more than 18000 images with meta-data including clothing category, and a \emph{manual shape annotation} indicating whether the person's shape is above average or average. The data comprises 181 different users. Using our multi-photo method, we estimated the shape of each user.
This allowed us to study the relationship between clothing categories and body shape. In particular, we compute the conditional distribution of clothing category conditioned on body shape parameters. 

To validate our conditional model, we compute the likelihood of the data (clothing categories worn by the user) under the model and compare it against multiple baselines, including a marginal model of clothing categories, 
a conditional model built using the manual shape annotations, and a conditional model using a state of the art single view shape estimation method~\cite{smplify}.

Experiments demonstrate that our conditional model with multi-photo shape estimates always produces better data-likelihood scores than the baselines. Notably, our model using automatic multi-photo shape estimation even outperforms a model using a coarse manual shape annotations. This shows that we extract more fine-grained shape information than manual annotations. This is remarkable, considering the unavoidable errors that sometimes occur in automatic shape estimation from monocular images. 

We see our method as the first step towards the large-scale understanding of clothing preferences from body shape. To stimulate further research in this direction, we will make the newly collected Fashion Takes Shape Dataset (FTS), and code available to the community. FTS includes clothing meta-data, 2D joint detection, semantic segmentation and our 3D shape-pose estimates.

\section{Related Work}
There is no previous work relating body shape to clothing preferences; here we review works that apply computer vision for fashion, and body shape estimation methods.

\textbf{Fashion Understanding in Computer Vision.}
Recently, fashion image understanding has gained a lot of attention in computer vision community, due to large range of its human-centric applications such as clothing recommendation~\cite{HipsterWarsECCV14,SimoSerraCVPR2015,liuLQWTcvpr16DeepFashion,han2017learning,Hsiao_2018_CVPR}, retrieval~\cite{Wang:2011:CSC:2072298.2072013,6751549,kiapour2015buy,Liu2012StreettoshopCC,Ak_2018_CVPR}, recognition~\cite{chen2012describing,liuLQWTcvpr16DeepFashion,Hsiao_2017_ICCV,han2017automatic,8237312}, parsing~\cite{6248101,yang2014clothing} and fashion landmark detection ~\cite{liuLQWTcvpr16DeepFashion,liuYLWTeccv16FashionLandmark,Wang_2018_CVPR}.

Whereas earlier work in this domain used handcrafted features (e.g. SIFT, HOG) to represent clothing~\cite{chen2012describing, HipsterWarsECCV14,Wang:2011:CSC:2072298.2072013,Liu2012StreettoshopCC}, newer approaches use deep learning~\cite{zhaobo_atman} which outperforms prior work by a large margin.
This is thanks to availability of large-scale fashion datasets~\cite{SimoSerraCVPR2015,liuLQWTcvpr16DeepFashion,liuYLWTeccv16FashionLandmark,han2017automatic,fashiondata} and blogs. Recent works in clothing recommendation leverage metadata from fashion blogs. In particular, Han et al.~\cite{han2017automatic} used the fashion collages to suggest outfits using multimodal user input.  Zhang et al.~\cite{7907314} studied the correlation between clothing and geographical locations and introduced a method to automatically recommend location-oriented clothing.
In another work, Simo-Serra et al.~\cite{SimoSerraCVPR2015} used user votes of fashion outfits to obtain a measure of fashionability.

Although the relationship between location, users vote and fashion compatibilities is well investigated, there is no work which studies the relationship between human body shape and clothing. In this work, we introduce an automatic method to estimate 3D shape and a model that relates it to clothing preferences. We also introduce a new dataset to promote further research in this direction.

\paragraph{Virtual try-on:} 
Another popular application of computer vision and computer graphics to fashion is virtual try-on, which boils down to a clothing re-targeting problem. Pons-Moll et al.~\cite{ponsmollSIGGRAPH17clothcap} jointly capture body shape and 3D clothing geometry -- which can be re-targeted to new bodies and poses. Other works by pass 3D inference;  using simple proxies for the body, Han et al.~\cite{han2017viton} retarget clothing to people directly in image space. 
Using deep learning and leveraging SMPL~\cite{SMPL:2015}, Lassner et al.~\cite{Lassner:GP:2017} predicts images of people in clothing, and Zanfir et al.~\cite{zanfir2018human} transfers appearance between subjects. These works leverage a body model to re-target clothing but do not study the correlation between body shape and clothing categories.

\paragraph{3D Body Shape Estimation.}
Recovery of 3D human shape from a 2D image is a very challenging task which has been facilitated by the availability of 3D generative body models learned from thousands of scans of people~\cite{anguelov2005scape,pons2015dyna,SMPL:2015}. Such models capture anthropometric constraints of the population and therefore reduce ambiguities. 
Several works~\cite{NIPS2007_3271,guan2009estimating,hasler2010multilinear,zhou2010parametric,10.1007/978-3-642-15558-1_22,smplify,MuVS:3DV:2017,hasler2010multilinear,zhou2010parametric,10.1007/978-3-642-15558-1_22} leverage these generative models to estimate 3D shape from single images using shading cues, silhouettes and appearance. 

Recent model based approaches leverage deep learning based 2D detections~\cite{cao2017realtime} -- by either fitting a model to them at test time~\cite{smplify,alldieck2018video} or by using them to supervise bottom-up 3D shape predictors~\cite{Pavlakos_2018_CVPR,Kanazawa_2018_CVPR,tung2017self,tan2017indirect}.
Similar to~\cite{smplify}, we fit the SMPL model to 2D joint detections, but, in order to obtain better shape estimates, we include a silhouette term in the objective like~\cite{alldieck2018video,MuVS:3DV:2017}.
In contrast to previous work, we leverage multiple web photos of the same person in different poses. In particular, we jointly optimize a single coherent static shape and pose for each of the images. This makes our multi-photo shape estimation approach robust to difficult poses and shape occluded by clothing. Other works have exploited temporal information in a sequence to estimate shape under clothing~\cite{bualan2008naked,shapeundercloth:CVPR17,yang2016estimation} in constrained settings -- in contrast we leverage web photos without known camera parameters. Furthermore, we can not assume pose coherency over time~\cite{MuVS:3DV:2017} since our input are photos with varied poses. None of previous work leverage multiple unconstrained photos of a person to estimate body shape.

\section{Robust Human Body Shape Estimation from Photo-Collections}\label{sec:smpl}
Our goal is to relate clothing preferences to body shape automatically inferred from photo-collections.
Here, we build on the SMPL~\cite{SMPL:2015} statistical body model that we fit to images. However, unconstrained online images make the problem very hard due to varying pose, clothing, illumination and depth ambiguities.  

To address these challenges, we propose a robust multi-photo estimation method. In contrast to controlled multi-view settings where the person is captured simultaneously by multiple cameras, we devise a method to estimate shape leveraging multiple photos of the same person in different poses and camera viewpoints.

From a collection of photos, our method starts by fitting SMPL (Sec.~\ref{sec:body_model}) to each of the images. This part is similar to~\cite{smplify,alldieck2018video} and not part of our contribution and we describe it in Sec.~\ref{sec:single_view} for completeness. We demonstrate (Sec.~\ref{sec:camera_optimization}) that keeping the height of the person fixed and initializing optimization at multiple depths significantly improves results and reduces scale ambiguities. Then, we reject photos that result in outlier shape estimates. Using the inlier photos, our multi-photo method (Sec.~\ref{sec:multi-photo}) jointly optimizes for multiple cameras, multiple poses, and a \emph{single coherent shape}. 
\subsection{Body Model}
\label{sec:body_model}
SMPL~\cite{SMPL:2015} is a state of the art generative body model, which parameterizes the surface of the human body with shape $\shape$ and pose $\pose$. The shape parameters $\shape \in \mathbb{R}^{10}$ are the PCA coefficients of a shape space learned from thousands of registered 3D scans. The shape parameters encode changes in height, weight and body proportions.
The body pose $\pose \in \mathbb{R}^{3P}$, is defined by a skeleton rig with $P = 24$ joints. The joints $J(\shape)$ are a function of shape parameters. 
The SMPL function $M(\shape, \pose)$ outputs the $N = 6890$ vertices of the human mesh transformed by pose $\pose$ and shape $\shape$.

In order to ``pose'' the 3D joints, SMPL applies global rigid transformations $\mathbf{R}_{\pose}$ on each joint $i$ as $R_{\pose}(J_i(\shape))$.

\subsection{Single View Fitting}
\label{sec:single_view}
We fit the SMPL model to 2D body joint detection $\mathbf{J}_{est}$ obtained using~\cite{insafutdinov2016deepercut}, and a foreground mask $\mathbf{S}$ computed using~\cite{SharpMask}. Concretely, we minimize an objective function with respect to pose, shape and camera translation $\mathbf{K}=[X,Y,Z]$:
\begin{multline}
E(\shape,\pose,\mathbf{K};\mathbf{J}_{\mathrm{est}},\mathbf{S})= E_P(\shape,\pose) + E_J(\shape,\pose; \mathbf{K},\mathbf{J}_{\mathrm{est}})\\
+E_h(\shape)+E_S(\shape,\pose,\mathbf{K};\mathbf{S}),
\label{eq:objective}
\end{multline}
where $E_P(\shape,\pose)$\footnote{For details please look at the supplementary material} are the four prior terms as described in~\cite{smplify}, and the other terms are described next. 

\paragraph{Joint-based data term:} We minimize the re-projection error between SMPL 3D joints and detection:
\begin{align}
E_J(\shape,\pose,\mathbf{K};\mathbf{J}_{\mathrm{est}})= \sum_{\mathrm{joint} j} \omega_j \rho(\Pi_{\mathbf{K}}(R_\theta(J(\shape)_j))-\mathbf{J}_{\mathrm{est},j})
\end{align} \label{eq:single}
where $\Pi_\mathbf{K}$ is the projection from 3D to 2D of the camera with parameters $\mathbf{K}$. $\omega_j$ are the confidence scores from CNN detection and $\rho$ a Geman-McClure penalty function which is robust to noise.

\paragraph{Height term:} Previous work~\cite{smplify,Lassner} jointly optimizes for depth (distance from the person to camera) and body shape. However, the overall size of the body and distance to the camera are ambiguous; a small person closer to the camera can produce a silhouette in the image just as big as a bigger person farther from the camera. 
Hence, we aim at estimating body shape up to a scale factor. To that end, we add an additional term that constrains the height of the person to remain very close to the mean height $T_H$ of the SMPL template
$$E_h(\shape) = ||M_h(0,\shape) - T_H||^2_2,$$
where height $M_h(0,\shape)$, is computed on the optimized SMPL model before applying pose. This step is especially crucial for multi-photo optimization as it allows us to analyze shapes at the same scale.

\paragraph{Silhouette term:} 
To capture shape better, we minimize the miss-match between the model silhouette $\mathbf{I_s}(\pose,\shape,\mathbf{K})$, and the distance transform of the CNN-segmented mask $\mathbf{S}$~\cite{SharpMask}:
\begin{equation}
E_S(\shape,\pose,\mathbf{K};\mathbf{S})=
 G(\lambda_{1}\mathbf{I_s}(\pose,\shape,\mathbf{K})\mathbf{S} + \lambda_{2}(\mathbf{1}-\mathbf{I_s}(\pose,\shape,\mathbf{K}))\overline{\mathbf{S}}),
\end{equation}
where $G$ is a Gaussian pyramid with 4 different levels, $\mathbf{K}$ is the camera parameter, and $\overline{\mathbf{S}}$ is the distance transform of the inverse segmentation, and $\lambda$ is a weight balancing the terms. 


\subsection{Camera Optimization}
\label{sec:camera_optimization}
Camera translation and body orientation are unknown. However, we assume that rough estimation of the focal length is known. We set the focal length as two times the width of the image. 
We initialize the depth $Z$ via the ratio of similar triangles, defined by the shoulder to ankle length of the 3D model and the 2D joint detection. To refine the estimated depth, we minimize the re-projection error $E_J$ of only torso, knee, and ankle joints (6 joints) with respect to camera translation and body root orientation.
At this stage, $\shape$ is held fixed to the template shape. 
We empirically found that a good depth initialization is crucial for good performance.
Hence, we minimize the objective in~\ref{eq:objective} at 5 different depth initializations -- we sample in the range of [-1,+1] meters from the initial depth estimate. We keep the shape estimate from the initialization that leads to a lower minimum after convergence. After obtaining the initial pose and shape parameter we refine the body shape model adding silhouette information.
\subsection{Robust Multi-Photo Optimization}
\label{sec:multi-photo}
The accuracy of the single view method heavily depends on the image view-point, the pose, the segmentation and 2D joint detection quality.
Therefore, we propose to jointly optimize one shape to fit several photos at once. Before optimizing, we reject photos that  are likely to be outliers in order to make the optimization more robust. First, we compute the median shape from all the single view estimates and keep only the views whose shape is closest to the median.
Using these inlier views we jointly optimize $V$ poses $\pose^i$, and a single shape $\shape$. We minimize the re-projection error in all the 
photos at once:
\begin{multline}
E_{mp}(\shape,\pose^{\forall i}, \mathbf{K}^{\forall i};\mathbf{J}^{\forall i}_{\mathrm{est}},\mathbf{S}^{\forall i})= \sum_{i=1}^{K} E(\shape,\pose^i,\mathbf{K}^i;\mathbf{J}^i_{\mathrm{est}},\mathbf{S}^i) \nonumber
\end{multline}
where $E()$ is the single view objective function, Eq.~\ref{eq:objective} and $K$ are number of views we kept after outlier rejection.
Our multi-photo method leads to more accurate shape estimates as we show in the experiments.

\section{Evaluation}
To evaluate our method, we proposed two datasets: synthetic and real images. 
We used the synthetic dataset to perform an ablation study of our multi-photo body model. 
Using unconstrained real-world fashion images, we evaluate our clothing category model conditioned on the multi-photo shape estimates.

\subsection{Synthetic Bodies}
\begin{figure}[h]
\centering
\begin{subfigure}[b]{0.39\textwidth}
\includegraphics[width=\linewidth]{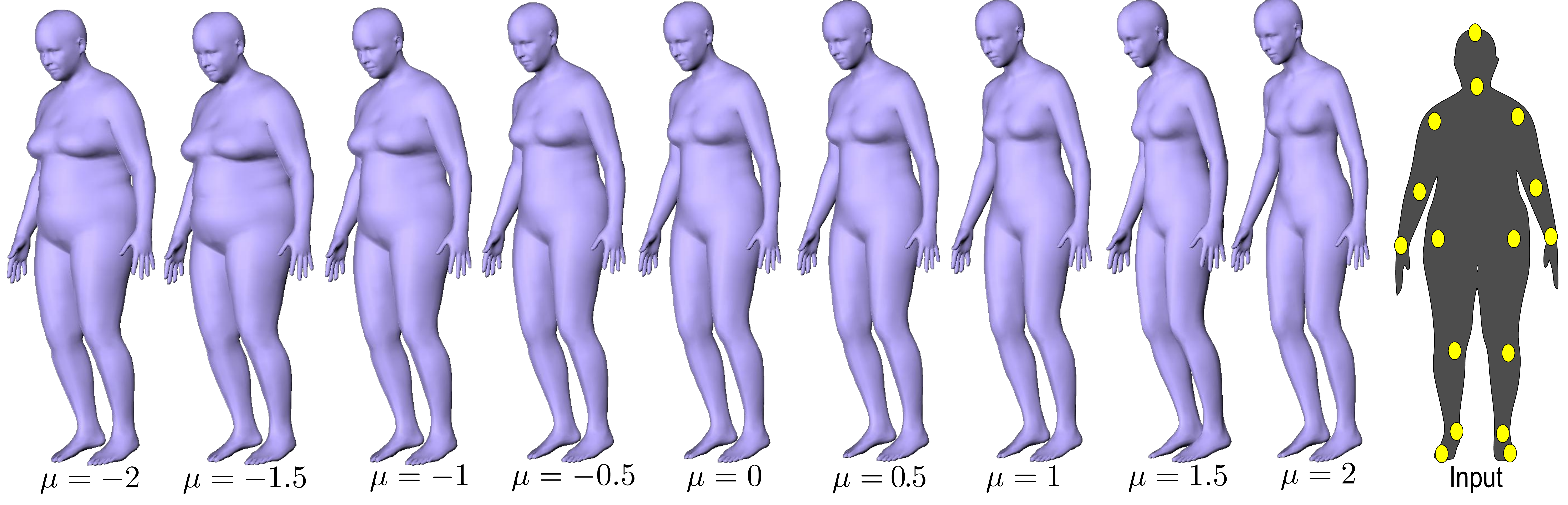}\par
        \caption{Variation in body shape}
        \label{fig:syn}
\end{subfigure}
\begin{subfigure}[b]{0.39\textwidth}
        \includegraphics[width=\textwidth]{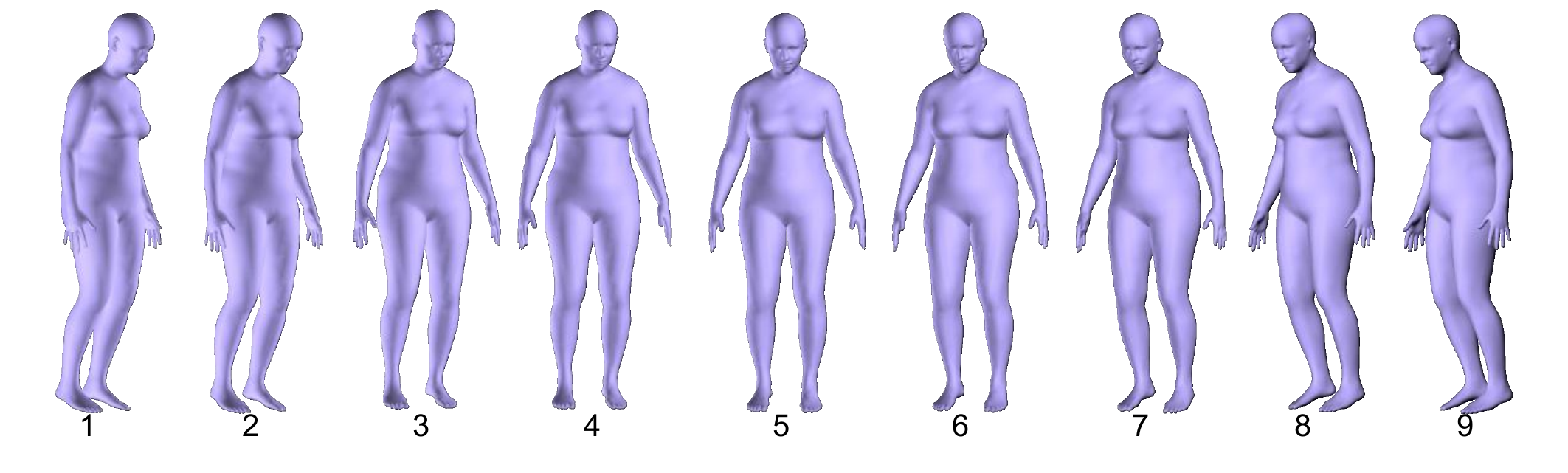}
        \caption{Variation in view}
        \label{fig:syn_view}
\end{subfigure}
 \caption{Using SMPL 3D body model we generate 9 subject each with 9 views. We study the effectiveness of our method on this dataset. As shown in (a) the input to our system is only image silhouette and a set of 2D body joints.}
\vspace*{\floatsep}
\end{figure}

SMPL is a generative 3D body model which is parametrized with pose and shape. We observe that while the first shape parameter produces body shape variations due to scale, the second parameter produces shapes of varied weight and form. Hence, we generate 9 different bodies by sampling the second shape parameter from $\mathcal{N}~(\mu,1)$ and $\mu \in [-2:0.5:2]$. In \autoref{fig:syn}, we show the 9 different body shapes and a representative rendered body silhouette with 2D joints which we use as input for prediction. We also generate 9 different views for each subject to evaluate our multi-photo shape estimation in a controlled setting (\autoref{fig:syn_view}). 
In all experiments, we report the mean Euclidean error between the estimated shape parameters $\shape$ and the ground truth shapes.

\vspace{-0.8cm}
\begin{figure}[h]
\centering
\includegraphics[width=0.7\linewidth]{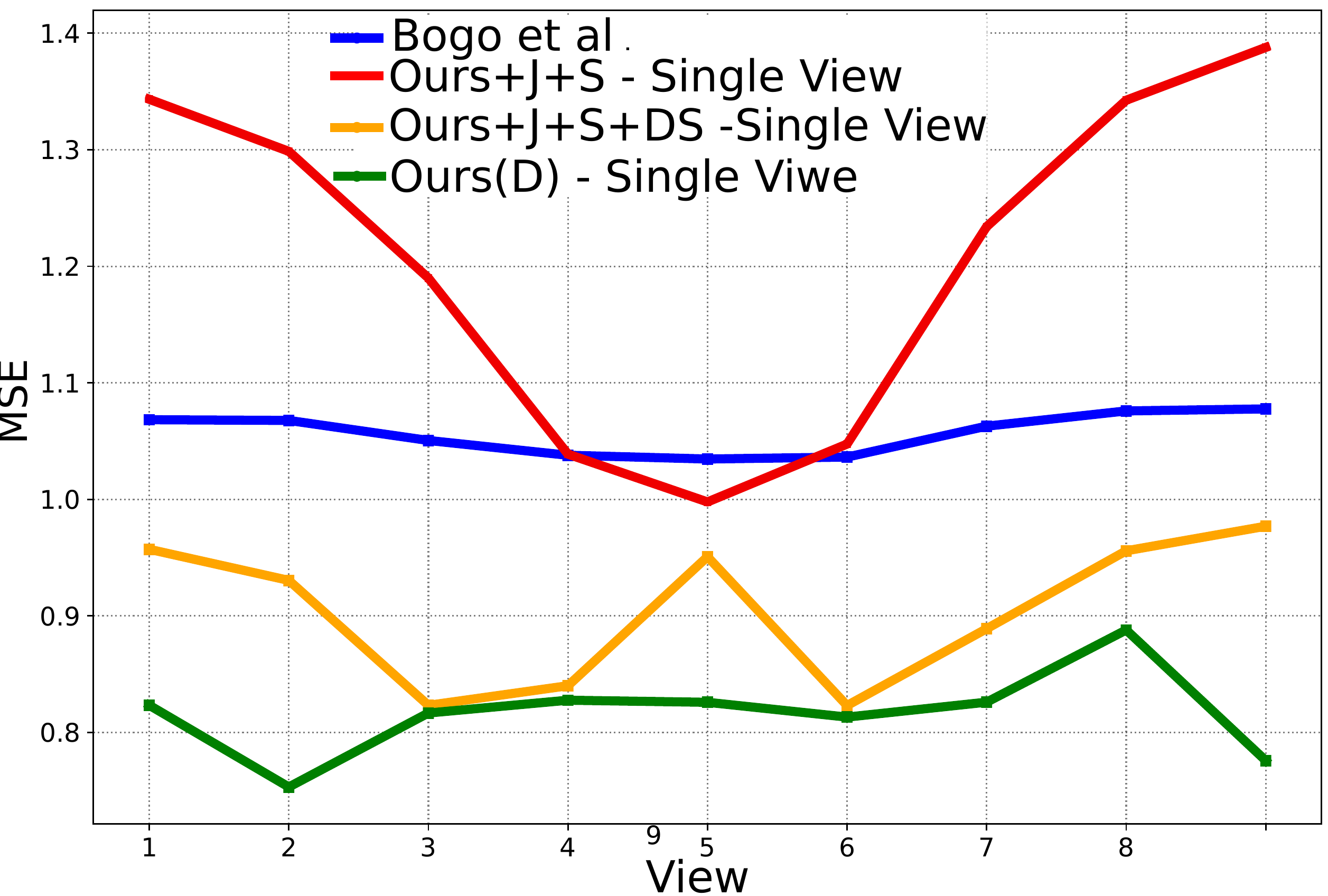}
\caption{The plot shows the mean euclidean norm between estimated and the ground truth shape among all subjects for each view on synthetic data.}\label{fig:res}
\end{figure}
\paragraph{Single-View Shape Estimation:}
In this controlled, synthetic setting, we have tested our model in several conditions. We summarize the results of our single view method in the first column of \autoref{tab:res} (mean shape error over all 9 subjects and 9 views), and plot the error w.r.t. viewpoints in~\autoref{fig:res}. Please note in~\autoref{tab:res} that SMPLify\cite{smplify} can only use one photo, and therefore columns corresponding to multiple photos are marked as ``na'' (not available). 

Overall, we see in \autoref{tab:res} a reduction of shape estimation error from 1.05 by SMPLify~\cite{smplify} to 0.91 of our method by adding joint estimation (J), silhouette features (S) and depth selection (DS). The depth selection (DS) strategy yield the strongest improvement. We see an additional decrease to 0.84 by considering multiple photos ($k=5$) despite having to estimate camera and body pose for each additional photo.

People with similar joint length could have different body mass. As SMPLify only uses 2D body joint as input, it is not able to estimate the body shape with high accuracy.
Silhouette is used to capture a better body shape. However adding silhouette with a wrong depth data, decreases the accuracy of shape estimation drastically (Ours using 2D joint and silhouette (Ours+J+S), red curve in \autoref{fig:res} from 0.91 to 1.20.
Hence, to study the impact of depth accuracy in shape estimation, we provide ground-truth depth (in~\autoref{fig:res}, green curve: \textit{Ours(D)}) to our method.
We observe that using ground truth depth information improves the error of Our+J+S from 1.20 to 0.86. 

This argues for the importance of our introduced depth search procedure. Indeed, we find that our model with depth selection (``Ours+J+S+DS'') yields a reduced error of 0.91 {\it without using any ground truth information.} 
 
\paragraph{Multi-Photo Shape Estimation:}
\autoref{tab:res} present also our result for multi-photo case. 
Real-world images exhibit noisy silhouettes and 2D joints, body occlusion, variation in camera parameters, viewpoints and poses. 
Consequently, we need a robust system that can use all the information to obtain an optimized shape. 
Since every single photo may not give us a very good shape estimation, we jointly optimize all photos together. 
However, for certain views, estimating the pose and depth is very difficult. Consequently, adding those views leads to worse performance. Hence, before optimization, we retain only the $K$ views with shape estimates closest to the median shape estimate of all views. This effectively rejects outlier views. 
The results are summarized in~\autoref{tab:res}. $K$ is the number of photos we kept out of the total of $9$ to perform optimization. 
Using only 2D joint data, we optimized the shape in multi-photo setting (Ours+J in~\autoref{tab:res}). In the second step we add silhouette term to our method in multi-photo optimization (Ours+J+S). Both of these experiments shows decrease in accuracy of the the estimated shape compared to SMPLify. 
However, for our full method which uses up to $K=5$ views, we observe a consistent decrease in error; beyond 5 the error increases, which supports the effectiveness of the proposed integration and outlier detection scheme. We improve over our single view estimate reducing the error {\it further} from 0.91 to 0.84 (for $K=5$) -- and even approach the oracle performance (0.86) where ground truth depth is given.
\begin{table}[!t]
\scalebox{0.75}{
\begin{tabular}{lllllllll}
\toprule
&single-view&k=2 &k=3 &k=4 &k=5 &k=6 & k=7 \\ \midrule
SMPLify~\cite{smplify}&1.05& na  & na & na& na & na& na \\
Ours+J&1.05& 1.81  & 1.80 & 1.80& 1.80 & 1.80& 1.82 \\
Ours+J+S &1.20&1.80 &1.80&1.80&1.80&1.80& 1.77\\
Ours+J+S+DS &0.91&0.91&0.88&0.87&0.84&0.85&0.88\\ \midrule
Ours(D)&0.86&0.84 &0.79&0.77&0.80&0.83&0.92\\\bottomrule
\end{tabular}}
\caption{We present the results for Multi-Photo optimization on the synthetic data. The error is the L2 distance between ground-truth and estimated shape parameter. Ours(D) has ground truth camera translation(depth) data.}
\label{tab:res}
\end{table}

\begin{figure}[h]
\centering
\includegraphics[width=0.7\linewidth, height=0.9\linewidth,keepaspectratio]{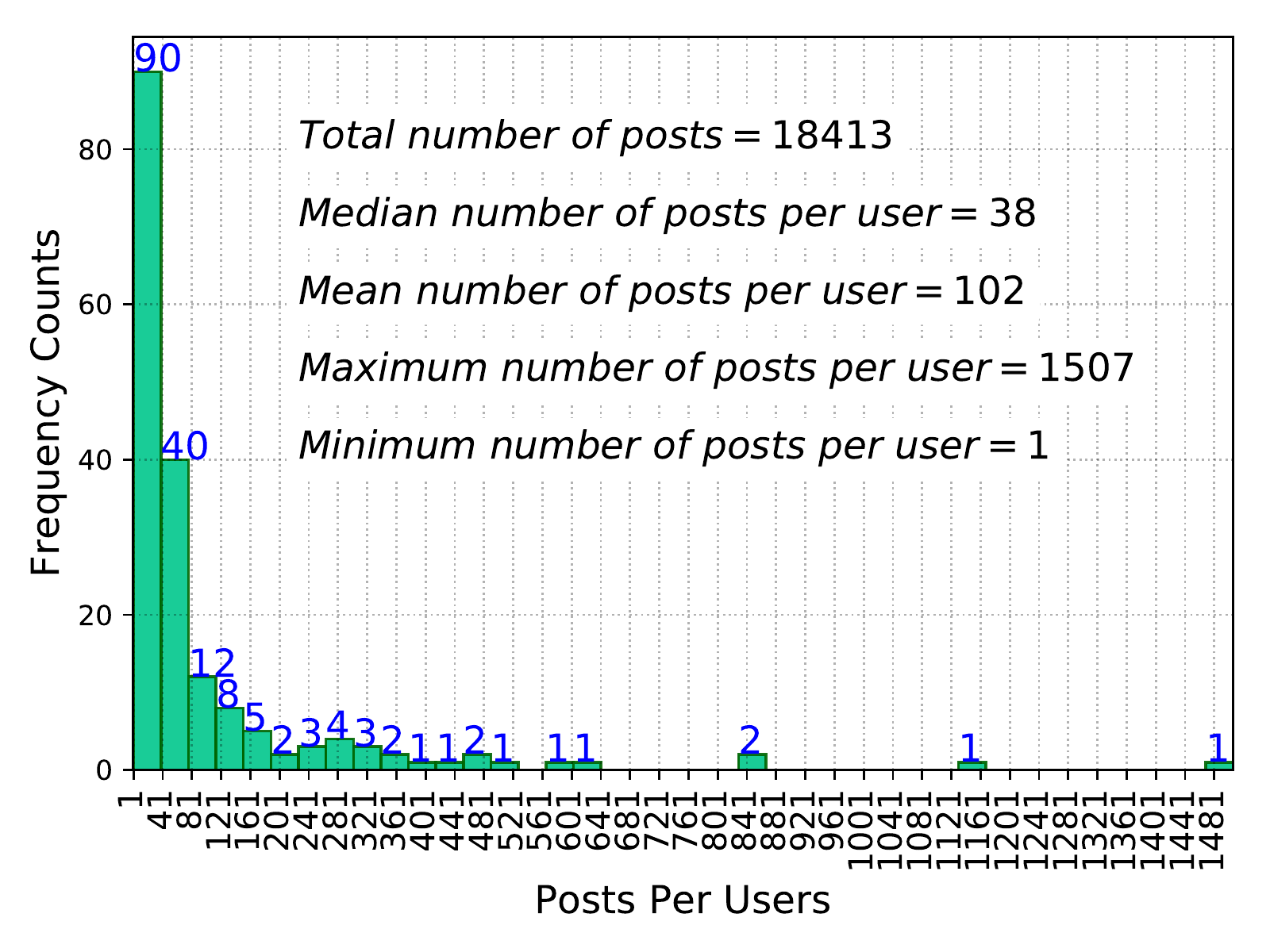}
\caption{Histogram of posts in our dataset. A total number of posts from all our users are 18413. Each post has 1 or more images of a person with clothing.}
\label{fig:data}
\end{figure}

\begin{figure}[h]
    \centering
  \includegraphics[width=0.65\linewidth, height=0.7\linewidth,keepaspectratio]{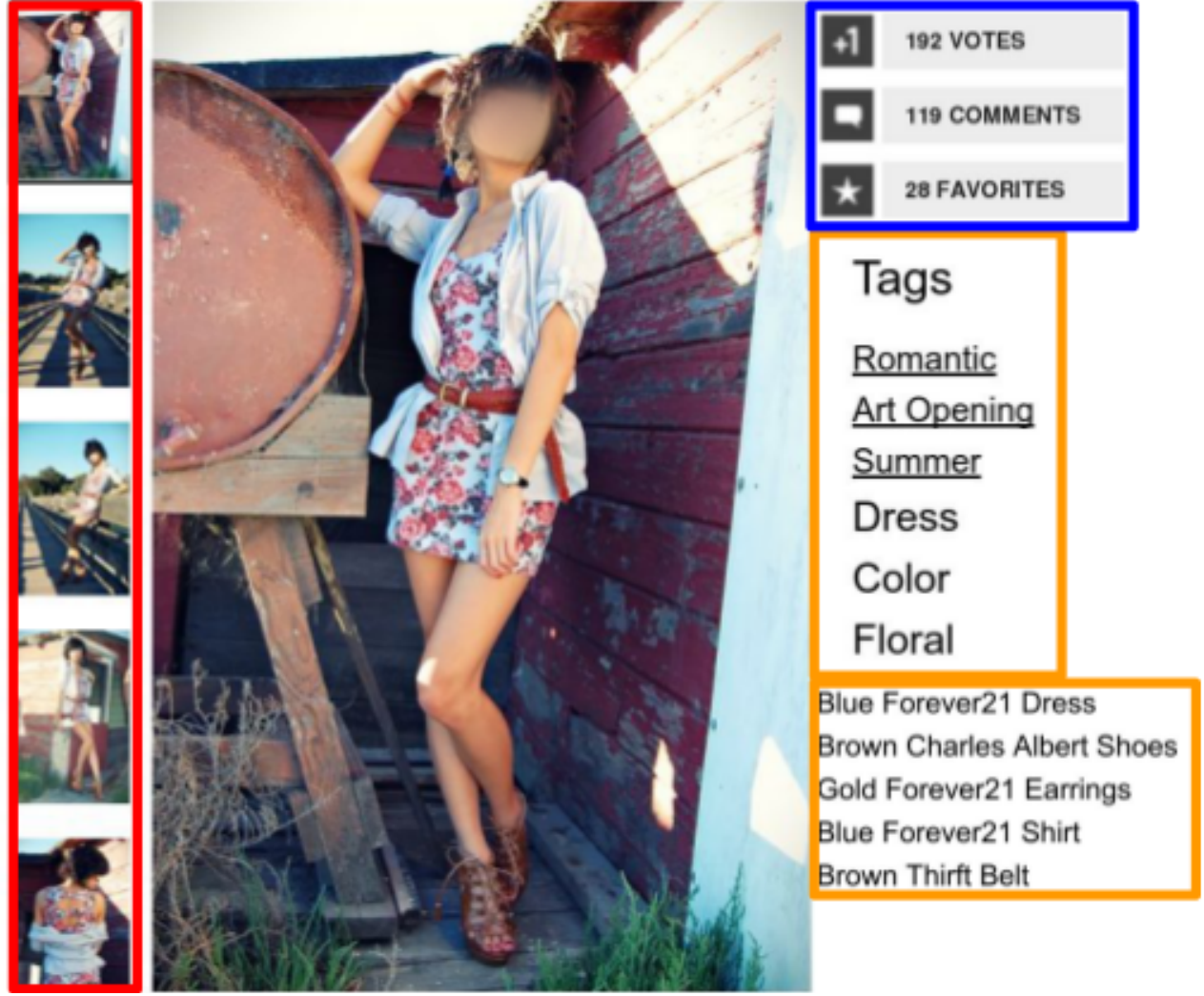}
  \caption{Chictopia's users upload images of themselves wearing different garments. Each post has 1 or more image of the person(red box). In addition to images, meta data such as "Tags"(orange box) and users opinion(blue box) are available.}
   \label{fig:post}
\end{figure}

\subsection{Fashion Takes Shape Dataset}
Not every clothing item matches every body shape. Hence our goal is to study the link between body shapes and clothing categories. In order to study these correlations, we collected data from 181 female users of ``Chictopia''\footnote{http://www.chictopia.com} (online fashion blog). 

We look for two sets of users: in the first set, we collected data of users with average and below the average size, which we call group $G_a$; the second set contains data of above average (plus size) users referred to as $G_p$.
In total, we have 141 users in group $G_a$ and 40 in group $G_p$ which constitute a diverse sample of real-world body shapes. 
\autoref{fig:data}, shows the summary of our dataset. The total number of posts from all users is 18413 -- each post can contains one or more images (usually between 2 to 4). 
The minimum number of posts per user is 1 and the maximum 1507. In average, we have 102 posts per user, and a median of 38 posts per user. Furthermore, each post contains data about clothing garments, other users opinions (Votes, likes) and comments.  \autoref{fig:post} shows a post uploaded by a user.  

\begin{figure}[h!]
\centering
\includegraphics[width=0.8\linewidth, height=0.9\linewidth,keepaspectratio]{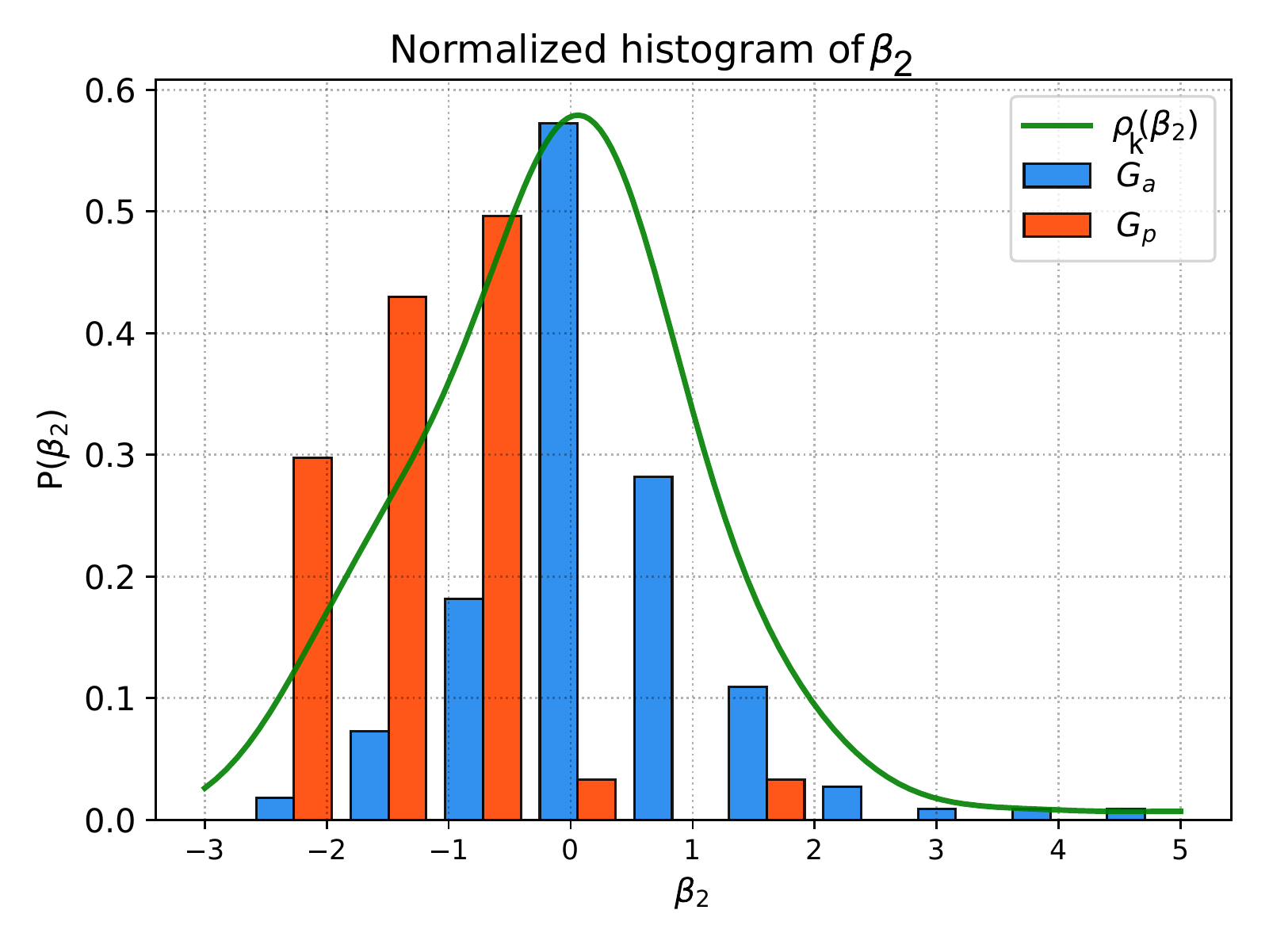}\par
\caption{Shape distribution of our dataset. The thinner the person the higher values of $\beta_2$ they have. While the group $G_p$ has lower values (negative).} \label{fig:p_betas}
\end{figure}

\subsection{Shape Representation}
In order to build a model conditioned on shape, we first need a representation of the users' shapes. Physical body measurements can be considered as an option which is not possible when we only have access to images of the person. 
Hence, we use our multi-photo method to obtain a shape $\beta \in \mathbb{R}^{10}$ estimate of the person from multiple photos.

Since we do not have ground truth shape for the unconstrained photos, we have trained a binary Support Vector Machine (SVM) on the estimated shape parameters $\beta$ for classification of the body type into $G_a$ and $G_p$. The intuition is that if our shape estimations are correct, above average and below average shapes should be separable. 
Indeed, the SVM obtains an accuracy (on a hold-out test set) of $87.17\%$ showing that the shape parameter is at the very least informative of the two aforementioned body classes.  
Looking at the SVM weights, we recognize that the second entry of the $\beta$ vector has the most contribution to the classifier. 
Actually, classifying the data by simple thresholding of the second dimension of the $\beta$ vector  results in an even higher accuracy of $88.79\%$. 
Hence, for following studies, we have used directly the second dimension of $\beta$. We illustrate the normalized histogram of this variable in \autoref{fig:p_betas}. The normalized histogram suggests that users in group $G_p$ have negative values whereas group $G_a$ have positive values. 
For later use, we estimate a probability density function (pdf) for this variable $\beta_2$ with a kernel density estimator -- using a Gaussian kernel: 
\begin{align}
\rho_K(\beta_2) = \frac{1}{N}\sum_{i=1}^{N} K((\beta_2 - \beta_2^{i}) / h)
\end{align}

The green line in \autoref{fig:p_betas} illustrate the estimated pdf $\rho_K(\beta_2)$ of all users. 

\subsection{Correlation Between Shape and Clothing Categories}
The type of clothing garment people wear is very closely correlated with their shape. For example, ``Leggings'' might look good on one body shape but may not look very good on other shapes. Hence, we introduce 3 models to study the correlation of the shape and clothing categories. Our basic model ({\bf Model 1}) uses only data statistics with no information about users shape. 
In the second approach ({\bf Model 2}), clothing is conditioned on binary shape categories $G_a$ and $G_p$ -- which in fact requires manual labels. The final approach ({\bf Model 3}) is facilitated by our automatic shape parameter estimation $\beta_2$. 

We evaluate the quality of the model via the negative log likelihood of held-out data. A good model minimizes the negative log-likelihood. Hence a better model should have smaller value in negative log likelihood.
The negative log likelihood is defined as: 
\begin{align}
LL =-\frac{1}{N}\sum_{i=1}^{N}\sum_{j=1}^{Z}\log P_{M}^{c^{i}_{j}}\
\end{align}
Where N is the number of users, Z is the number of clothing categories and $c^{i}_{j} \in {C_i}$ is a vector of user's clothing categories.  $M$ represents the model. 
In the following, we present the details of these three approaches. The log likelihood of each approach is reported in the \autoref{tab:best}.

\begin{figure}[t]
    \centering
  \includegraphics[width=0.9\linewidth]{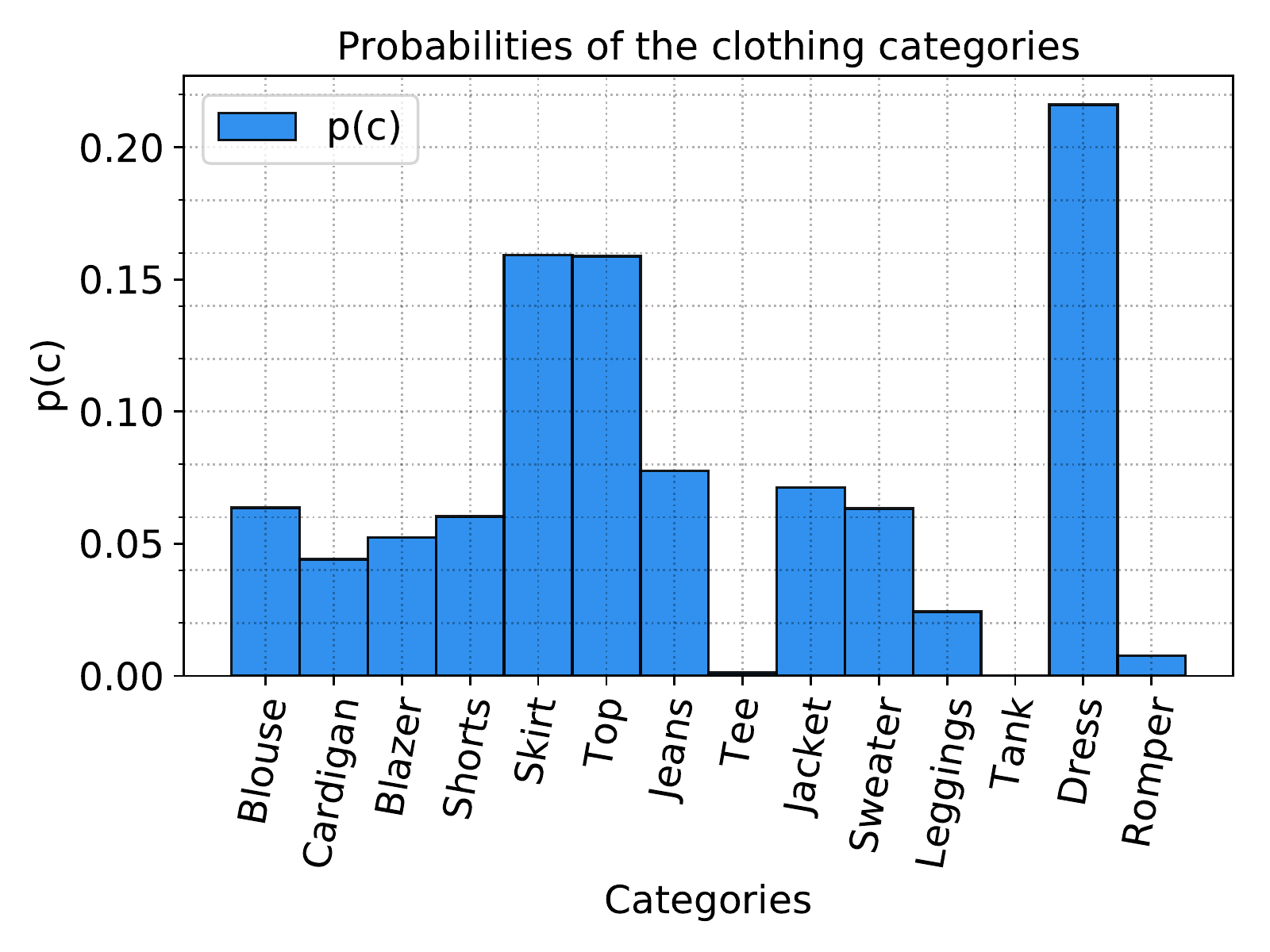}
  \caption{Probabilities of 14 different clothing garments of our dataset.}
  \label{fig:pc}
\end{figure}

\begin{figure}[h]
\centering 
\includegraphics[width=0.9\linewidth]{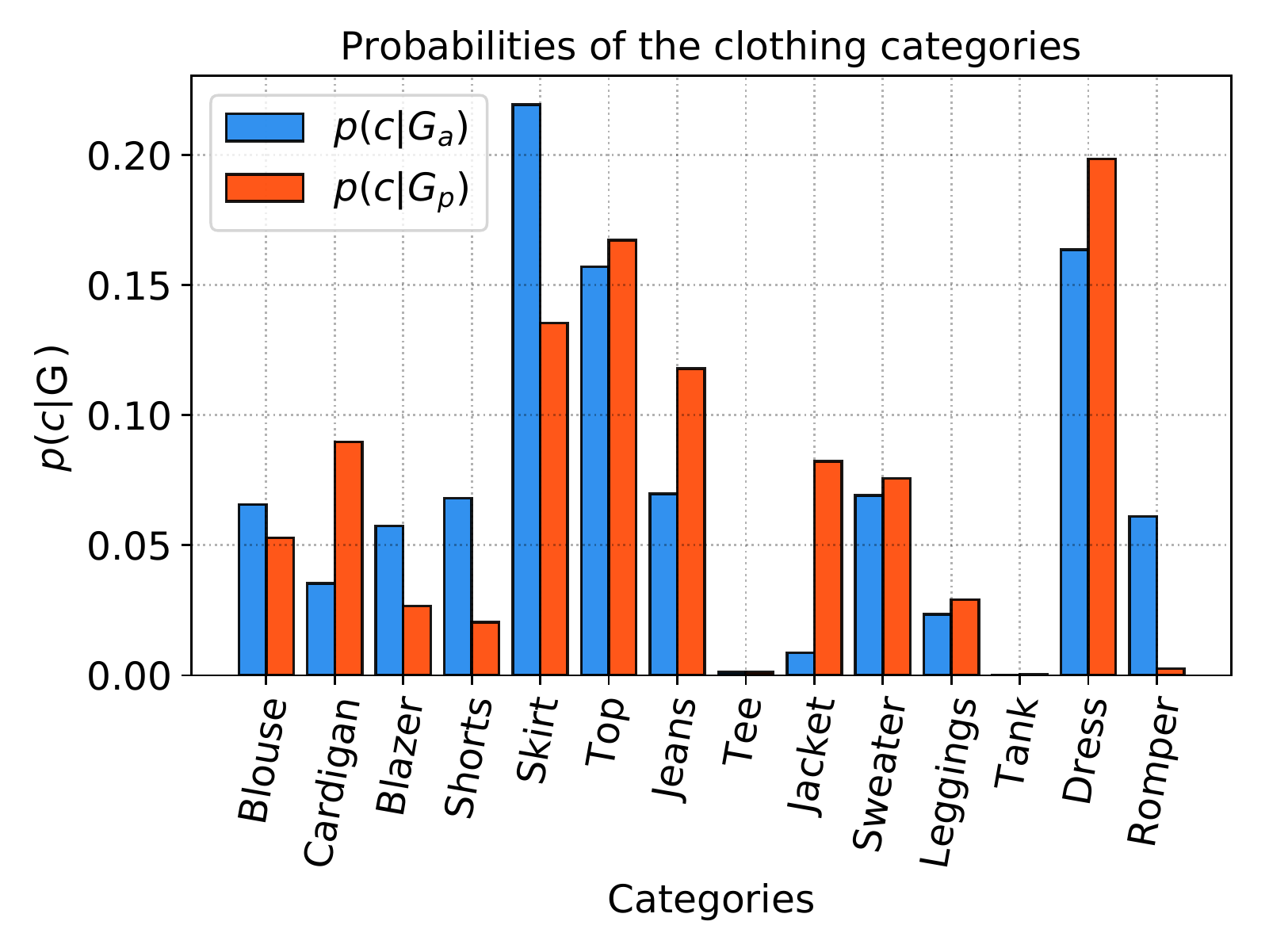}\par
 \caption{Given the Body type ($G_a$ and $G_p$), we measured the probability of each clothing category in our dataset.}
 \label{fig:p_n}
\end{figure}

\paragraph{Model 1: Prediction Using Probability of Clothing Categories:}
We established a basic model using the probability of the clothing categories $P_M = p(c)$. 
The clothing categories tag of the dataset is parsed for fourteen of the most common clothing's categories. Category ``Dress'' has the highest amount of images (\autoref{fig:pc}) whereas ``Tee and Tank'' were not tagged very often. 
\vspace{-0.5cm}
\paragraph{Model 2: Prediction for Given $G_a$ and $G_p$:}\label{sec:p_n}
This  is based on the conditional probability of the clothing category given the annotated body type $P_M = p(c|G)$ where $G\in {G_a, G_p}$. From this measure we find out that several clothing categories are more likely for certain group (\autoref{fig:p_n}).  
As an example, while ``Cardigan'' and ``Jacket'' have higher probabilities for the $G_p$ group, users in $G_a$ were more likely to wear ``Short'' and ``Skirt''.
\begin{figure*}[ht]
\vspace{-1cm}
\centering
 \includegraphics[width=0.95\textwidth]{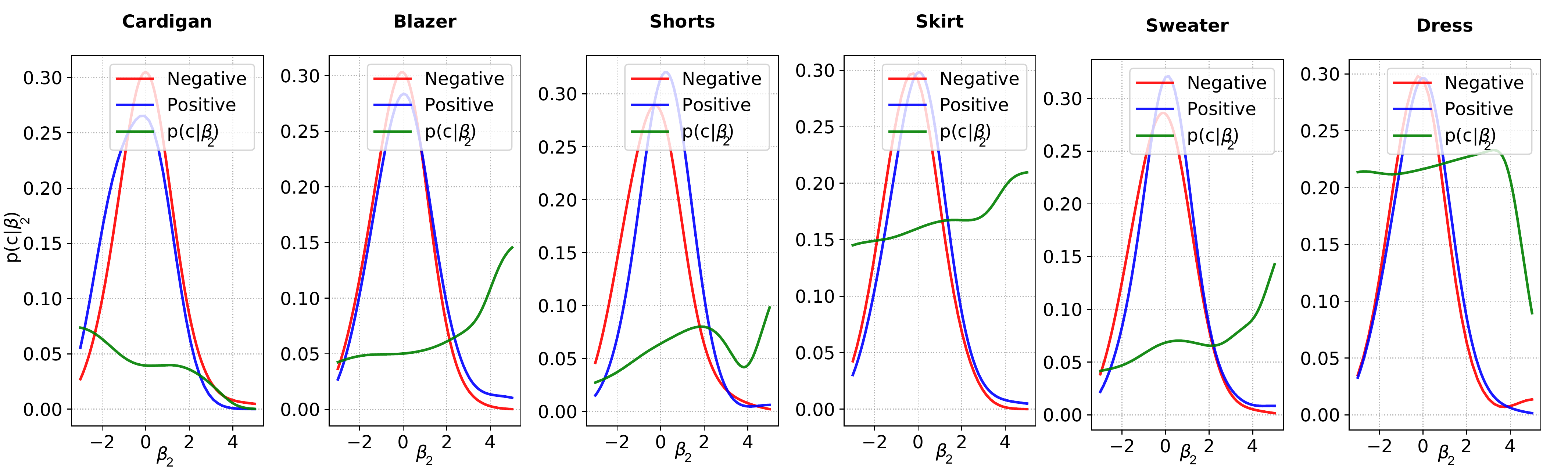}
\caption{Probabilities of $p(\beta_2|c)$ for wearing (blue curves) and not wearing (red curves) of a clothing category on our dataset. Using Bayes rule we can estimate the probability of clothing given the shape $p(c|\beta_2)$(green curve). Negative values of $\beta_2$ corresponds to above average while average and below average users have positive values for $\beta_2$.
\label{p_b}}
\end{figure*}
\paragraph{Model 3: Prediction for a Given Shape  $\beta_2$:}
As shown in the second model, body types and clothing garments are correlated. However, categorizing people only into two or more categories is not desirable. First, it requires tedious and time consuming manual annotation of body type. 
Second, the definition of the shape categories is very personal and fuzzy. 

The estimated shape parameters of our model provides us with a continuous fine-grained representation of human body shape. Hence, we no further need to classify people in arbitrary shape groups. 
Using the shape parameter $\beta_2$ and statistics of our data, we are able to measure the conditional probability of shapes for a given clothing category $p(\beta_2|c)$. This probability is measured for wearing and not wearing a certain category. The result is shown in \autoref{p_b} where for each category the Blue line represents wearing the category. 
Similarly to the previous model, one can see users with negative values of $\beta_2$ wearing ``Cardigan'', where the probabilities of wearing ``Short'' and ``skirt'' is skewed towards positive values of $\beta_2$.
Furthermore, using the Bayes rule we can predict clothing condition on the body shape $P_M =p(c|\beta_2)$ as: \begin{align}
p(c|\beta_2)=\frac{p(\beta_2|c)p(c)}{P(\beta_2)}
\end{align}
The green line in \autoref{p_b} illustrate the $p(c|\beta_2)$. 

\paragraph{Negative Log likelihood} We quantify the quality of our prediction models by the negative log likelihood of held out data. As we are using negative of log likelihood, the model with smallest values is the best. 
The results, of each model on our dataset, is summarized in the \autoref{tab:best}.
Also, we used the estimated shape parameters of Bogo et al~\cite{smplify} and ours for comparison. In addition to our method which optimized multi-photo, we only used the median shape among photos of a user as a baseline as well. We also include the likelihood under the prior as a reference.
For better analysis, we split the users into 4 groups: The first group contains users which there is only a single photo ($I = 1$) of them in each post. In the second group, users always have 2 images($I = 2$) and third group contains users with 3 or more images of a clothing($I\geq3$). Finally, we also show results with taking into account all users. 
\autoref{tab:best}, shows that our method obtained the smallest negative log-likelihood on the full dataset -- in particular outperforming the model that conditions on the two discrete labeled shape classes, shape based on prior work SMPLify\cite{smplify}, as well as a naive multi-photo integration based on a median estimate. While the median estimate is comparable if only two views are available, we see significant gains for multiple viewpoint -- that also show on the full dataset.

\begin{table}[!h]
\begin{center}
\scalebox{0.75}{
\begin{tabular}{@{}lcccccc@{}}\\
\toprule
& Model 1 & Model 2 & & & Model 3\\
 & p(c) & p(c$|G_a$/$G_p$)& $p(c|\beta_2)^1$ & Median$\beta_2$ & $p(c|\beta_2)^2$\\\midrule 
I = 1 & 12.81& 12.80 & 13.63 & - &\bf{12.16} \\
I = 2 & 13.31& 13.47 & 13.34 & \bf{13.09} &13.11 \\
I $\geq$ 3 &19.06 &19.11&18.8& 18.59&\bf{17.85} \\\midrule
All & 20.13& 20.39 & 20.48 & 20.12&\bf{19.81} \\
\bottomrule
\end{tabular}}
\end{center}
\caption{We measured the negative Log-Likelihood of our different models on held out data. Numbers are comparable within the rows. Smaller is better. $p(c|\beta_2)^1$ uses estimated shape from SMPLify\cite{smplify} and $p(c|\beta_2)^2$ uses our estimated shape.}
\label{tab:best}
\end{table}

\section{Qualitative Results on Shape Estimation}
\begin{figure}[h]
 \centering
\includegraphics[width=0.9\linewidth,keepaspectratio]{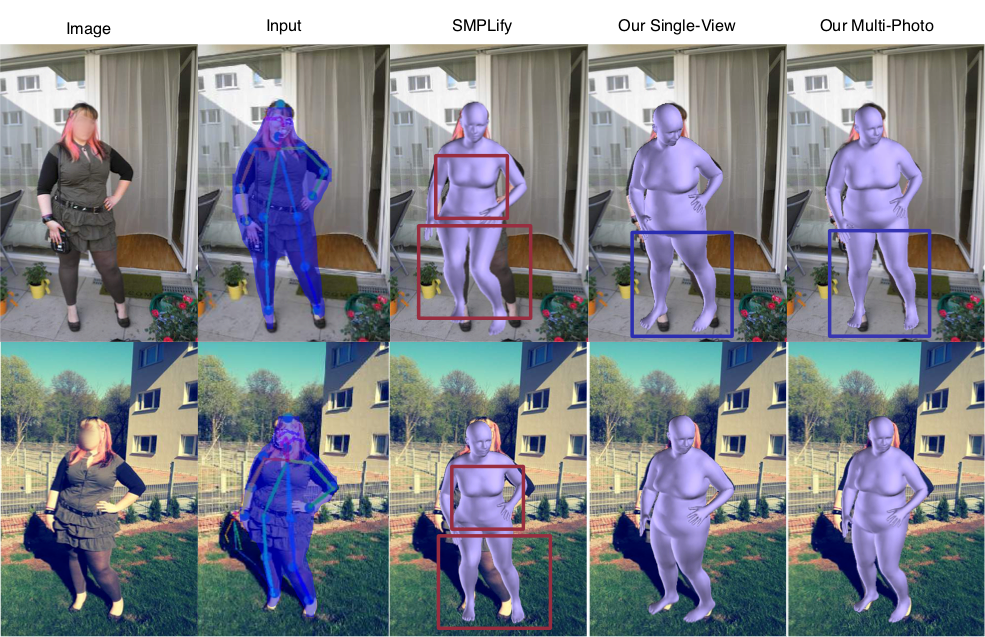}\\
\includegraphics[width=0.9\linewidth,keepaspectratio]{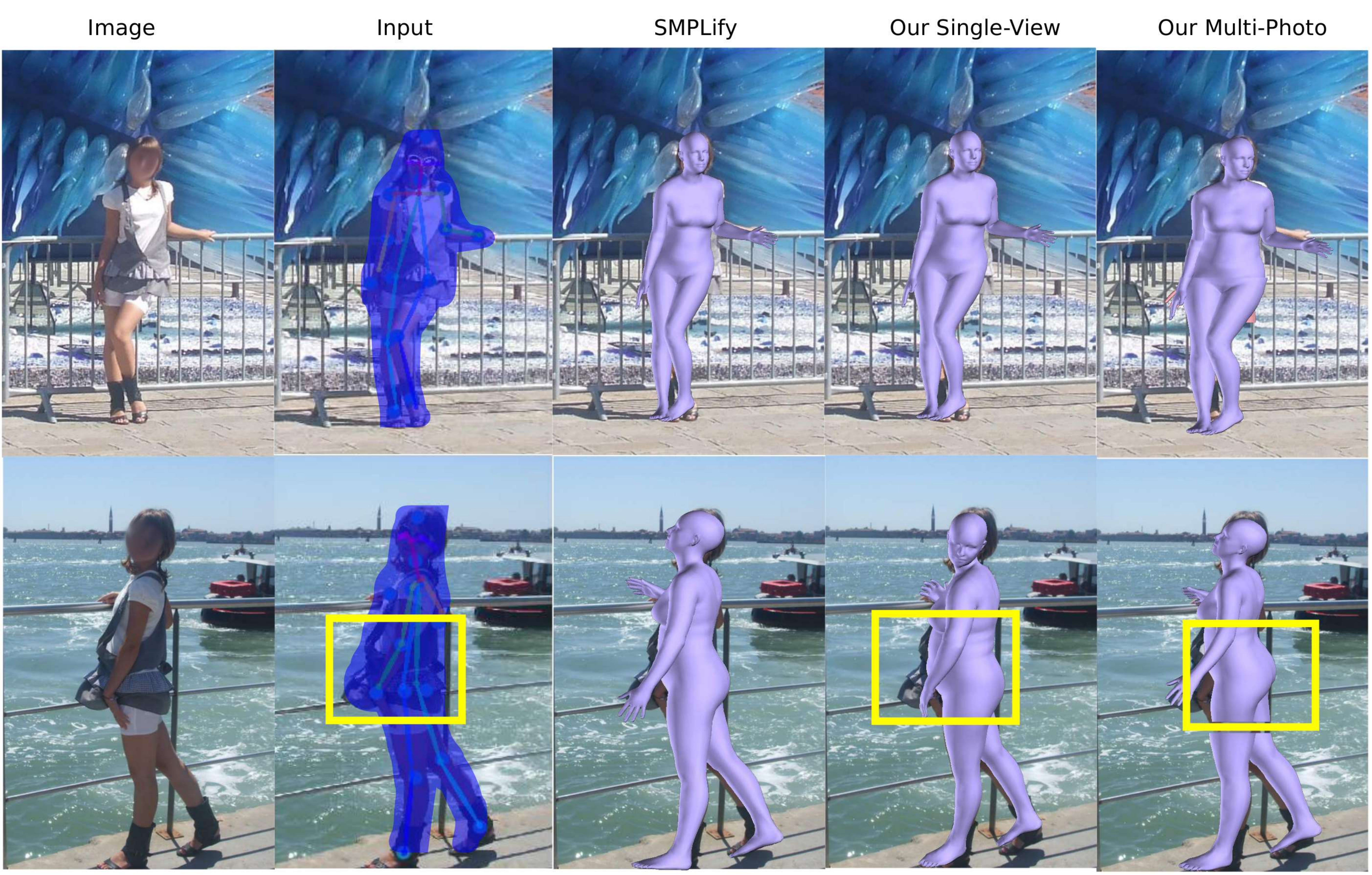}
\caption{Shape estimation results on real data. Note that the shape estimates obtained with SMPLify~\cite{smplify} are rather close to the average body shape whereas our multi-photo approach is able to recover shape details more accurately both for above-average (rows 1 and 2) and average (rows 3 and 4) body types.}\label{fig:res_real}
\end{figure}
In \autoref{fig:res_real} we present example results obtained with our method and compare it to the result obtained with SMPLify~\cite{smplify}. SMPLify fits the body model based on 2D positions of body joints that often do not provide enough information regarding body girth. This leads to shape estimates that are rather close to the average body shape for above-average body sizes (rows 1 and 2 in \autoref{fig:res_real}). SMLPify also occasionally fails to select the correct depth that results in body shape that is too tall and has bent knees (red box). The single-view variant of our approach improves over the result of SIMPLify for the first example in \autoref{fig:res_real}. However it still fails to estimate the fine-grained pose details such as orientation of the feet (blue box). In the second example in \autoref{fig:res_real} the body segmentation includes a handbag resulting in a shape estimate with exaggerated girth by our single-view approach (yellow box). These mistakes are corrected by our multi-photo approach that is able to improve feet orientation in the first example (blue box) and body shape in the second example (yellow box).
\section{Conclusion}
In this paper we aimed to understand the connection between body shapes and clothing preferences by collecting and analyzing a large database of fashion photographs with annotated clothing categories. Our results demonstrate that clothing preferences and body shapes are correlated and that we can build predictive models for clothing categories based on the output of automatic shape estimation. To obtain estimates of 3D shape we proposed a new approach that incorporates evidence from multiple photographs and body segmentation and is more accurate than popular recent SMPLify approach ~\cite{smplify}. We are making our data and code available for research purposes.
\newpage
\clearpage
\newpage

{\small
\bibliographystyle{ieee}
\bibliography{egbib}
}
\end{document}